\documentclass[runningheads,a4paper]{llncs}

\usepackage{amssymb}
\usepackage{amsmath}
\usepackage{graphicx}
\usepackage{multirow}
\usepackage[normalem]{ulem}
\useunder{\uline}{\ul}{}
\usepackage{bbding}

\usepackage{pbox}
\usepackage{makecell}
\usepackage{ifxetex}
\usepackage{ifluatex}

\ifxetex
  \usepackage[ios,font=seguiemj.ttf]{emoji}
  \usepackage{fontspec}
  
\else
  \ifluatex
    \usepackage[ios,font=Symbola_hint.ttf]{emoji}
    \usepackage{fontspec}
    
  \else
    \usepackage[T1]{fontenc}
    \usepackage[utf8]{inputenc}
    \usepackage[ios]{emoji}
  \fi
\fi

\usepackage{url}
\urldef{\mailsa}\path|{amit, sujan, sanjaya, tkprasad}@knoesis.org|
\newcommand{\keywords}[1]{\par\addvspace\baselineskip
\noindent\keywordname\enspace\ignorespaces#1}

\begin{document}

\mainmatter  

\pdfinfo{
/Title (Knowledge will Propel Machine Understanding of Content: Extrapolating from Current Examples)
/Author (Amit Sheth,Sujan Perera,Sanjaya Wijeratne, Krishnaprasad Thirunarayan)
/Keywords (Semantic Analysis of Multimodal Data;Knowledge-enabled Computing;Machine Intelligence;Multimodal Exploitation;Understanding Complex Text;Knowledge-enhanced ML and NLP;Knowledge-driven Deep Content Understanding)
/Subject (Knowledge will Propel Machine Understanding of Content: Extrapolating from Current Examples)}

\title{Knowledge will Propel Machine Understanding of Content: Extrapolating from Current Examples}

\titlerunning{Knowledge will Propel Machine Understanding of Content}

\author{Amit Sheth \and Sujan Perera \and Sanjaya Wijeratne \and Krishnaprasad Thirunarayan}
\authorrunning{Sheth et al.}

\institute{Kno.e.sis Center, Wright State University\\
Dayton, Ohio, USA\\
\mailsa\\
\url{http://www.knoesis.org}}

\maketitle

\begin{abstract}

Machine Learning has been a big success story during the AI resurgence. One particular stand out success relates to learning from a massive amount of data. In spite of early assertions of the unreasonable effectiveness of data, there is increasing recognition for utilizing knowledge whenever it is available or can be created purposefully. In this paper, we discuss the indispensable role of knowledge for deeper understanding of content where (i) large amounts of training data are unavailable, (ii) the objects to be recognized are complex, (e.g., implicit entities and highly subjective content), and (iii) applications need to use complementary or related data in multiple modalities/media. What brings us to the cusp of rapid progress is our ability to (a) create relevant and reliable knowledge and (b) carefully exploit knowledge to enhance ML/NLP techniques. Using diverse examples, we seek to foretell unprecedented progress in our ability for deeper understanding and exploitation of multimodal data and continued incorporation of knowledge in learning techniques.


\end{abstract}

\section{Introduction} 

Recent success in the area of Machine Learning (ML) for Natural Language Processing (NLP) has been largely credited to the availability of enormous training datasets and computing power to train complex computational models~\cite{halevy2009unreasonable}. Complex NLP tasks such as statistical machine translation and speech recognition have greatly benefited from the Web-scale unlabeled data that is freely available for consumption by learning systems such as deep neural nets. However, many traditional research problems related to NLP, such as part-of-speech tagging and named entity recognition (NER), require labeled or human-annotated data, but the creation of such datasets is expensive in terms of the human effort required. In spite of early assertion of the unreasonable effectiveness of data (i.e., data alone is sufficient), there is an increasing recognition for utilizing knowledge to solve complex AI problems. Even though knowledge base creation and curation is non-trivial, it can significantly improve result quality, reliability, and coverage. A number of AI experts, including Yoav Shoham~\cite{shoham2015knowledge}, Oren Etzioni, and Pedro Domingos~\cite{domingos2012few,domingos2015master}, have talked about this in recent years. In fact, codification and exploitation of declarative knowledge can be both feasible and beneficial in situations where there is not enough data or adequate methodology to learn the nuances associated with  the concepts and their relationships.

The value of domain/world knowledge in solving complex problems was recognized much earlier~\cite{winograd1972understanding}. These efforts were centered around language understanding. Hence, the major focus was towards representing linguistic knowledge. The most popular artifacts of these efforts are FrameNet~\cite{ruppenhofer2006framenet} and WordNet~\cite{miller1990introduction}, which were developed by realizing the ideas of frame semantics~\cite{fillmore1976frame} and lexical-semantic relations~\cite{cruse1986lexical}, respectively. Both these resources have been used extensively by the NLP research community to understand the semantics of natural language documents.

The building and utilization of the knowledge bases took a major leap with the advent of the Semantic Web in the early 2000s. For example, it was the key to the first patent on Semantic Web and a commercial semantic search/browsing and personalization engine over 15 years ago~\cite{sheth2001system}, where knowledge in multiple domains complemented ML techniques for information extraction (NER, semantic annotation) and  building intelligent applications\footnote{\url{http://j.mp/15yrsSS}}. Major efforts in the Semantic Web community have produced large, cross-domain (e.g., DBpedia, Yago, Freebase, Google Knowledge Graph) and domain specific (e.g., Gene Ontology, MusicBrainz, UMLS) knowledge bases in recent years which have served as the foundation for the intelligent applications discussed next.

The value of these knowledge bases has been demonstrated for determining semantic similarity~\cite{meng2013review,emojisimilarity}, question answering~\cite{shekarpour2013question}, ontology alignment~\cite{jain2010ontology}, and word sense disambiguation (WSD)~\cite{mihalcea2006knowledge}, as well as major practical AI services, including Apple's Siri, Google's Semantic Search, and IBM's Watson. For example, Siri relies on knowledge extracted from reputed online resources to answer queries on  restaurant searches, movie suggestions, nearby events, etc. In fact, ``question answering'', which is the core competency of Siri, was built by partnering with Semantic Web and Semantic Search service providers who extensively utilize knowledge bases in their applications\footnote{\url{https://en.wikipedia.org/wiki/Siri}}. The Jeopardy version of IBM Watson uses semi-structured and structured knowledge bases such as DBpedia, Yago, and WordNet to strengthen the evidence and answer sources to fuel its DeepQA architecture~\cite{ferrucci2010building}. A recent study~\cite{mcmahon2017substantial} has shown that Google search results can be negatively affected when it does not have access to Wikipedia. Google Semantic Search is fueled by Google Knowledge Graph\footnote{\url{http://bit.ly/22xUjZ6}}, which is also used to enrich search results similar to what the Taalee/Semagix semantic search engine did 15 years ago\footnote{\url{https://goo.gl/A54hno}}~\cite{sheth2001system,sheth2002managing}.

While knowledge bases are used in an auxiliary manner in the above scenarios, we argue that they have a major role to play in understanding real-world data. Real-world data has a greater complexity that has yet to be fully appreciated and supported by automated systems. This complexity emerges from various dimensions. Human communication has added many constructs to language that help people better organize knowledge and communicate effectively and concisely. However, current information extraction solutions fall short in processing several implicit constructs and information that is readily accessible to humans. One source of such complexity is our ability to express ideas, facts, and opinions in an implicit manner. For example, the sentence \textit{``The patient showed accumulation of fluid in his extremities, but respirations were unlabored and there were no use of accessory muscles''} refers to the clinical conditions of ``shortness of breath'' and ``edema'', which would be understood by a clinician. However, the sentence does not contain names of these clinical conditions -- rather it contains descriptions that imply the two conditions. Current literature on entity extraction has not paid much attention to implicit entities~\cite{rizzo2015making}.

Another complexity in real-world scenarios and use cases is data heterogeneity due to their multimodal nature. There is an increasing availability of physical (including sensor/IoT), cyber, and social data that are related to events and experiences of human interest~\cite{sheth2013physical}. For example, in our personalized digital health application for managing asthma in children\footnote{\url{http://bit.ly/kAsthma}}, we use numeric data from sensors for measuring a patient's physiology (e.g., exhaled nitric oxide) and immediate surroundings (e.g., volatile organic compounds, particulate matter, temperature, humidity), collect data from the Web for the local area (e.g., air quality, pollen, weather), and extract textual data from social media (i.e., tweets and web forum data relevant to asthma)~\cite{anantharam2015knowledge}. Each of these modalities provides complementary information that is helpful in evaluating a hypothesis provided by a clinician and also helps in disease management. We can also relate anomalies in the sensor readings (such as spirometer) to asthma symptoms and potential treatments (such as taking rescue medication). Thus, understanding a patient's health and well-being requires integrating and interpreting multimodal data and gleaning insights to provide reliable situational awareness and decisions. Knowledge bases play a critical role in establishing relationships between multiple data streams of diverse modalities, disease characteristics and treatments, and in transcending multiple abstraction levels~\cite{sheth2016semanticcognitive}. For instance, we can relate the asthma severity level of a patient, measured exhaled nitric oxide, relevant environmental triggers, and prescribed asthma medications to one another to come up with personalized actionable insights and decisions.

Knowledge bases can come in handy when there is not enough hand-labaled data for supervised learning. For example, emoji sense disambiguation, which is the ability to identify the meaning of an emoji in the context of a message in a computational manner~\cite{emojinet,emojineticwsm}, is a problem that can be solved using supervised and knowledge-based approaches. However, there is no hand-labeled emoji sense dataset in existence that can be used to solve this problem using supervised learning algorithms. One reason for this could be that emoji have only recently become popular, despite having been first introduced in the late 1990s~\cite{emojinet}. We have developed a comprehensive emoji sense knowledge base called EmojiNet~\cite{emojinet,emojineticwsm} by automatically extracting emoji senses from open web resources and integrating them with BabelNet. Using EmojiNet as a sense inventory, we have demonstrated that the emoji sense disambiguation problem can be solved with carefully designed knowledge bases, obtaining promising results~\cite{emojineticwsm}.

In this paper, we argue that careful exploitation of knowledge can greatly enhance the current ability of (big) data processing. At Kno.e.sis, we have dealt with several complex situations where:

\begin{enumerate}
    \item Large quantities of hand-labeled data required for unsupervised (self-taught) techniques to work well is not available or the annotation effort is significant.
    \item The text to be recognized is complex (i.e., beyond simple entity - person/location/organization), requiring novel techniques for dealing with complex/compound entities~\cite{ramakrishnan2008joint}, implicit entities~\cite{perera2015implicit,perera2016implicit}, and subjectivity (emotions, intention)~\cite{jadhav2016knowledge,wang2015automatic}.
    \item Multimodal data -- numeric, textual and image, qualitative and quantitative, certain and uncertain -- are available naturally~\cite{anantharam2015knowledge,anantharam2016understanding,lakshikagang,gangwordembeddings}.
\end{enumerate}

Our recent efforts have centered around exploiting different kinds of knowledge bases and using semantic techniques to complement and enhance ML, statistical techniques, and NLP. Our ideas are inspired by the human brain's ability to learn and generalize knowledge from a small amount of data (i.e., humans do not need to examine tens of thousands of cat faces to recognize the next ``unseen'' cat shown to them), analyze situations by simultaneously and synergistically exploiting multimodal data streams, and understand more complex and nuanced aspects of content, especially by knowing (through common-sense knowledge) semantics/identity preserving transformations.

\section{Challenges in creating and using knowledge bases}
Last decade saw an increasing use of background knowledge for solving diverse problems. While applications such as searching, browsing, and question answering can use large, publically available knowledge bases in their current form, others like movie recommendation, biomedical knowledge discovery, and clinical data interpretation are challenged by the limitations discussed below. \\

\noindent 
\textbf{Lack of organization of knowledge bases:} 
Proper organization of knowledge bases has not kept pace with their rapid growth, both in terms of variety and size. Users find it increasingly difficult to find relevant knowledge bases or relevant portions of a large knowledge base for use in domain-specific applications (e.g., movie, clinical, biomedical). This highlights the need to identify and select relevant knowledge bases such as the linked open data cloud, and extract the relevant portion of the knowledge from broad coverage sources such as Wikipedia and DBpedia. We are working on automatically indexing the domains of the knowledge bases~\cite{lalithsena2013automatic} and exploiting the semantics of the entities and their relationships to select relevant portions of a knowledge base \cite{lalithsena2016harnessing}.\\

\noindent 
\textbf{Gaps in represented knowledge:} 
The existing knowledge bases can be incomplete with respect to a task at hand. For example, applications such as computer assisted coding (CAC) and clinical document improvement (CDI) require comprehensive knowledge about a particular domain (e.g., cardiology, oncology)\footnote{\url{https://goo.gl/nXDY8x}}. We observe that although the existing medical knowledge bases (e.g., Unified Medical Language System (UMLS)) are rich in taxonomical relationships, they lack non-taxonomical relationships among clinical entities. We have developed data-driven algorithms that use real-world clinical data (such as EMRs) to discover missing relationships between clinical entities in existing knowledge base, and then get these validated by a domain-expert-in-the-loop~\cite{perera2014semantics}. Yet another challenge is creating personalized knowledge bases for specific tasks. For example, in~\cite{sheth2016semantic}, personal knowledge graphs are created based on the content consumed by a user, taking into account the dynamically changing vocabulary, and this is applied to improve subsequent filtering of relevant content.\\

\noindent 
\textbf{Inefficient metadata representation and reasoning techniques:} 
The scope of what is captured in the knowledge bases is rapidly expanding, and involves capturing more subtle aspects such as subjectivity (intention, emotions, sentiments), spatial and temporal information, and provenance. Traditional triple-based representation languages developed by Semantic Web community (e.g., RDF, OWL) are unsuitable for capturing such metadata due to their limited expressivity. For example, representation of spatio-temporal context or uncertainty associated with a triple is {\it ad hoc\/}, inefficient, and lacks semantic integration for formal reasoning. These limitations and requirements are well-recognized by the Semantic Web community, with some recent promising research to address them. For example, the singleton-property based representation~\cite{nguyen2014don} adds ability to make statements about a triple (i.e., to express the context of a triple) and probabilistic soft logic~\cite{kimmig2012short} adds ability to associate the probability value with a triple and reason over them. It will be really exciting to see applications exploiting such enhanced hybrid knowledge representation models that perform `human-like' reasoning on them.

Next, we discuss several applications that utilize knowledge bases and multimodal data to circumvent or overcoming some of the aforementioned challenges due to insufficient manually-created knowledge. \\

\noindent \textbf{Application 1: Emoji sense disambiguation}

\begin{figure*}
\centering
\includegraphics[width=1.0\linewidth]{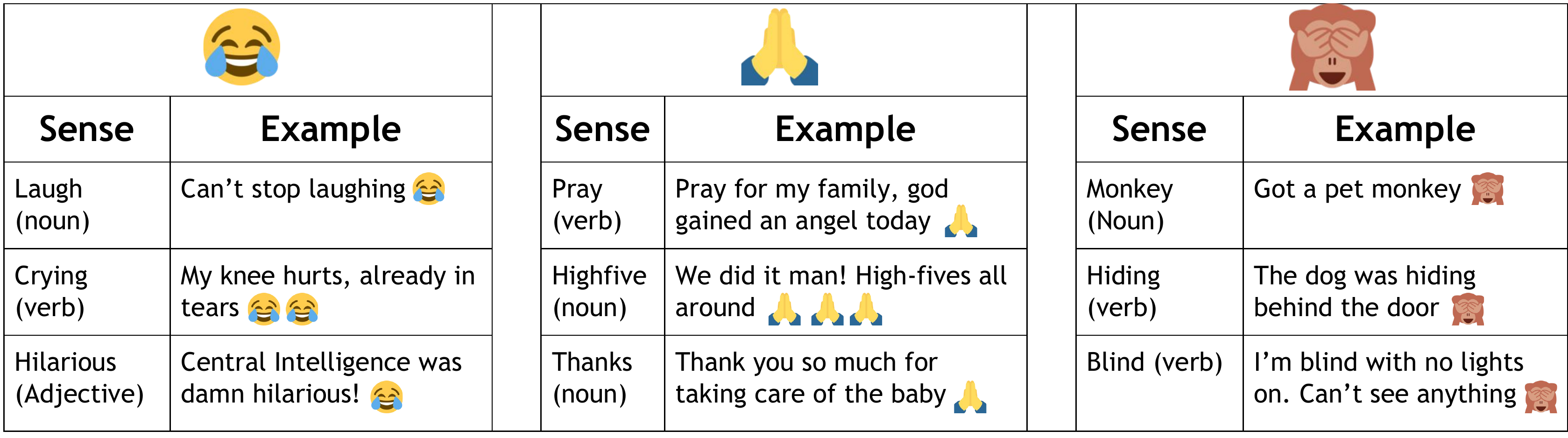} 
\caption{Emoji usage in social media with multiple senses.}
\label{emojiensesexamples}
\end{figure*}

With the rise of social media, ``emoji'' have become extremely popular in online communication. People are using emoji as a new language on social media to add color and whimsiness to their messages. Without rigid semantics attached to them, emoji symbols take on different meanings based on the context of a message. This has resulted in ambiguity in emoji use (see Figure~\ref{emojiensesexamples}). Only recently have there been efforts to mimic NLP techniques used for machine translation, word sense disambiguation and search into the realm of emoji~\cite{emojineticwsm}. The ability to automatically process, derive meaning, and interpret text fused with emoji will be essential as society embraces emoji as a standard form of online communication. Having access to knowledge bases that are specifically designed to capture emoji meaning can play a vital role in representing, contextually disambiguating, and converting pictorial forms of emoji into text, thereby leveraging and generalizing NLP techniques for processing richer medium of communication. 

As a step towards building machines that can understand emoji, we have developed EmojiNet~\cite{emojinet,emojineticwsm}, the first machine readable sense inventory for emoji. It links Unicode emoji representations to their English meanings extracted from the Web, enabling systems to link emoji with their context-specific meanings. EmojiNet is constructed by integrating multiple emoji resources with BabelNet, which is the most comprehensive multilingual sense inventory available to-date. For example, for the emoji `face with tears of joy' \emoji{1F602}, EmojiNet lists 14 different senses, ranging from happy to sad. An application designed to disambiguate emoji senses can use the senses provided by EmojiNet to automatically learn message contexts where a particular emoji sense could appear. Emoji sense disambiguation could improve the research on sentiment and emotion analysis. For example, consider the emoji \emoji{1F602}, which can take the meanings \textit{happy} and \textit{sad} based on the context in which it has been used. Current sentiment analysis applications do not differentiate among these two meanings when they process \emoji{1F602}. However, finding the meanings of \emoji{1F602} by emoji sense disambiguation techniques~\cite{emojineticwsm} can improve sentiment prediction. Emoji similarity calculation is another task that could be benefited by knowledge bases and multi-modal data analysis. Similar to computing similarity between words, we can calculate the similarity between emoji characters. We have demonstrated how EmojiNet can be utilized to solve the problem of emoji similarity~\cite{emojisimilarity}. Specifically, we have shown that emoji similarity measures based on the rich emoji meanings available in EmojiNet can outperform conventional emoji similarity measures based on distributional semantic models and also helps to improve applications such as sentiment analysis~\cite{emojisimilarity}. \\

 \noindent \textbf{Application 2: Implicit entity linking}
 
As discussed, one of the complexities in data is the ability to express facts, ideas, and opinions in an implicit manner. As humans, we seamlessly express and infer implicit information in our daily conversations. Consider the two tweets \textit{``Aren't we gonna talk about how ridiculous the new space movie with Sandra Bullock is?''} and \textit{``I'm striving to be +ve in what I say, so I'll refrain from making a comment abt the latest Michael Bay movie''}. The first tweet contains an implicit mention of movie `Gravity' and the second tweet contains an element of sarcasm and negative sentiment towards the movie `Transformers: Age of Extinction'. Both the sentiment and the movie are implicit in the tweet. While it is possible to express facts, ideas, and opinions in an implicit manner, for brevity, we will focus on how knowledge aids in automatic identification of implicitly mentioned entities in text.

We define implicit entities as ``entities mentioned in text where neither its name nor its synonym/alias/abbreviation or co-reference is explicitly mentioned in the same text''. 
Implicit entities are a common occurrence. For example, our studies found that 21\% of movie mentions and 40\% of book mentions are implicit in tweets, and about 35\% and 40\% of `edema' and `shortness of breath' mentions are implicit in clinical narratives. There are genuine reasons why people tend to use implicit mentions in daily conversations. Here are few reasons that we have observed:

\begin{enumerate}
\item To express sentiment and sarcasm : See above examples.
\item To provide descriptive information : For example, it is a common practice to describe the features of an entity rather than simply list down its name in clinical narratives. Consider the sentence `small fluid adjacent to the gallbladder with gallstones which may represent inflammation.' This sentence contains implicit mention of the condition cholecystitis (`inflammation in gallbladder' is recognized as cholecystitis) with its possible cause. The extra information (i.e., possible cause) in description can be critical in understanding the patient's health status and treating the patient. While it is feasible to provide these extra information with the corresponding explicit entity names, it is observed that clinical professionals prefer this style.
\item To emphasize the features of an entity : Sometimes we replace the name of the entity with its special characteristics in order to give importance to those characteristics. For example, the text snippet ``Mason Evans 12 year long shoot won big in golden globe'' has an implicit mention of the movie `Boyhood.' There is a difference between this text snippet and its alternative form ``Boyhood won big in golden globe.'' The speaker is interested in emphasizing the distinct feature of the movie, which would have been ignored if he had used the name of the movie as in the second phrase.
\item To communicate shared understanding : We do not bother spelling out everything when we know that the other person has enough background knowledge to understand the message conveyed. A good example is the fact that clinical narratives rarely mention the relationships between entities explicitly (e.g., relationships between symptoms and disorders, relationships between medications and disorders), rather it is understood that the other professionals reading the document have the expertise to understand such implicit relationships in the document.
\end{enumerate}

\begin{figure}
\centering
\includegraphics[scale=0.35]{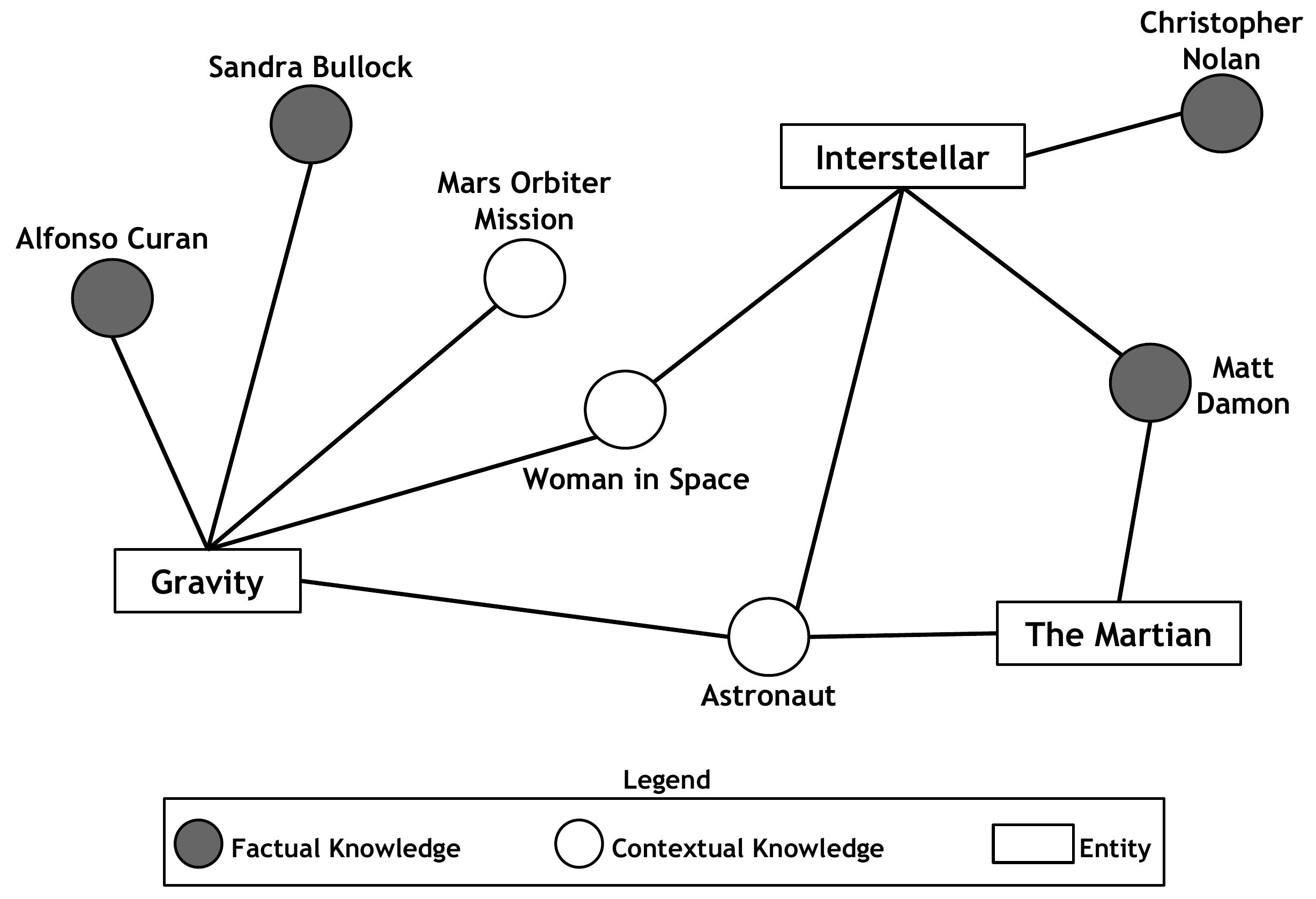}
\caption{Entity model extracted for three movies.}
\label{implicit}
\end{figure}

Whenever we communicate, we assume common understanding or shared-knowledge with the audience. A reader who does not know that Sandra Bullock starred in the movie `Gravity' and that it is a space exploration movie would not be able to decode the reference to the movie `Gravity' in the first example; a reader who does not know about Michael Bay's movie release would have no clue about the movie mentioned in the second tweet; a reader who does not know the characteristics of the clinical condition 'cholecystitis' would not be able to decode its mention in the clinical text snippet shown above; a reader who is not a medical expert would not be able to connect the diseases and symptoms mentioned in a clinical narrative. \textbf{These examples demonstrate the indispensable value of domain knowledge in text understanding}. Unfortunately, state-of-the-art named entity recognition applications do not capture implicit entities~\cite{rizzo2015making}. Also, we have not seen big data-centric or other approaches that can glean implicit entities without the use of background knowledge (that is already available (e.g., in UMLS) or can be created (e.g., from tweets and Wikipedia)).

The task of recognizing implicit entities in text demands comprehensive and up-to-date world knowledge. Individuals resort to a diverse set of entity characteristics to make implicit references. For example, references to the movie `Boyhood' can use phrases like \textit{``Richard Linklater movie''}, \textit{``Ellar Coltrane on his 12-year movie role''}, \textit{``12-year long movie shoot''}, \textit{``latest movie shot in my city Houston''}, and \textit{``Mason Evan's childhood movie''}. Hence, it is important to have comprehensive knowledge about the entities to decode their implicit mentions. Another complexity is the temporal relevancy of the knowledge. The same phrase can be used to refer to different entities at different points in time. For instance, the phrase \textit{``space movie''}  referred to the movie `Gravity' in Fall 2013, while the same phrase in Fall 2015 referred to the movie `The Martian'. On the flip side, the most salient characteristics of a movie may change over time and so will the phrases used to refer to it. In November 2014 the movie `Furious 7' was frequently referred to with the phrase \textit{``Paul Walker's last movie''}. This was due to the actor's death around that time. However, after the movie release in April 2015, the same movie was often mentioned through the phrase \textit{``fastest film to reach the \$1 billion''}.

We have developed knowledge-driven solutions that decode the implicit entity mentions in clinical narratives~\cite{perera2015implicit} and tweets~\cite{perera2016implicit}. We exploit the publicly available knowledge bases (only the portions that matches with the domain of interest) in order to access the required domain knowledge to decode implicitly mentioned entities. Our solution models individual entities of interest by collecting knowledge about the entities from these publicly available knowledge bases, which consist of definitions of the entities, other associated concepts, and the temporal relevance of the associated concepts. Figure~\ref{implicit} shows a snippet from generated entity model. It shows the models generated for movies `Gravity', `Interstellar', and `The Martian'. The colored (shaded) nodes (circles) represent factual knowledge related to these movies extracted from DBpedia knowledge base and the uncolored nodes represent the contextual knowledge (time-sensitive knowledge) related to entities extracted from daily communications in Twitter. The implicit entity linking algorithms are designed to carefully use the knowledge encoded in these models to identify implicit entities in the text.\\

\begin{figure*}
\centering
\includegraphics[scale=0.16]{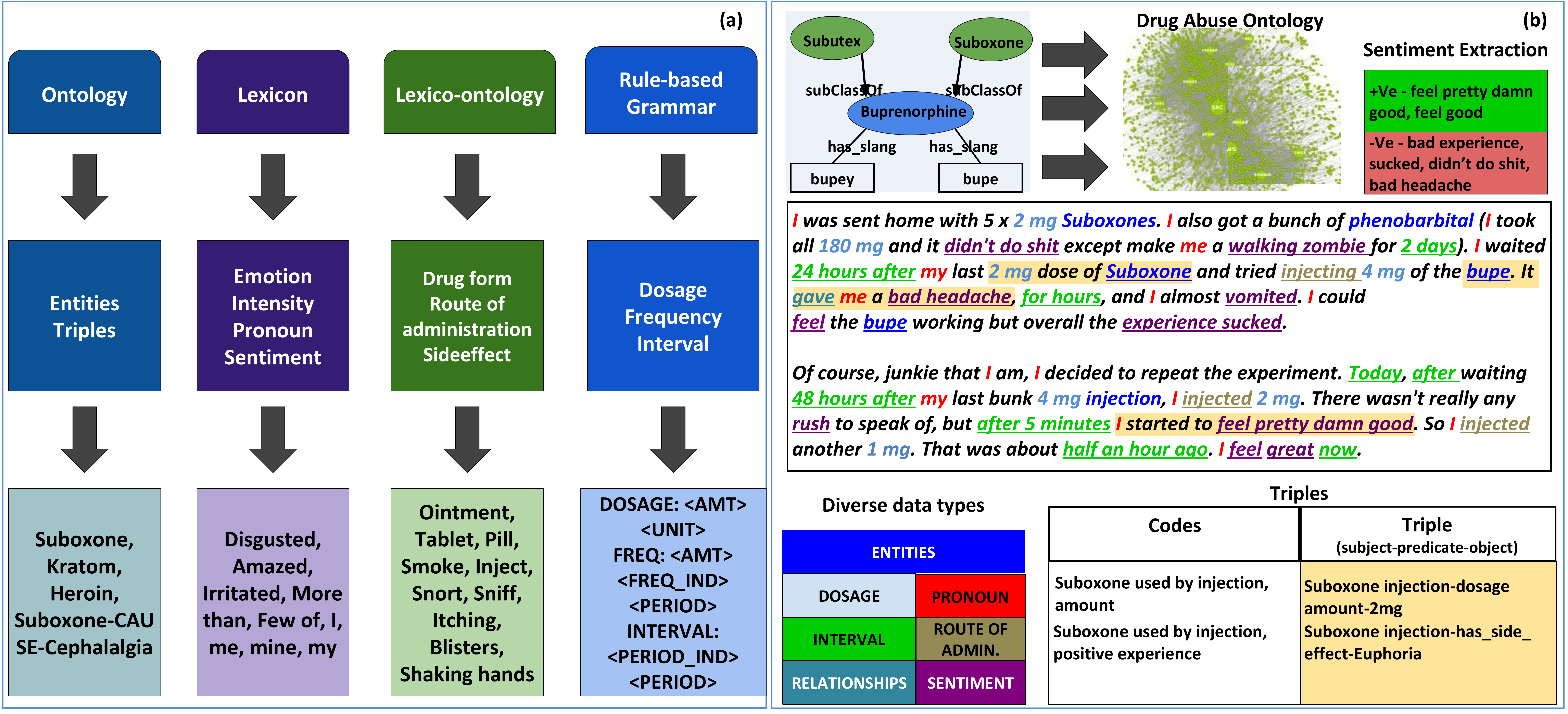}
\caption{(a) Use of background knowledge to enhance information extraction of diverse types of information. (b) Example use of diverse knowledge and information extraction for deeper and more comprehensive understanding of text in health and drug abuse domain. See~\cite{cameron2014hybrid} for more information.}
\label{fig:fbgusers}
\end{figure*}

\noindent \textbf{Application 3: Understanding and analyzing drug abuse related discussions on web forums}

The use of knowledge bases to improve keyword-based search has received much attention from commercial search engines lately. However, the use of knowledge bases alone cannot solve complex, domain-specific information needs. For example, answering a complex search query such as ``How are drug users overdosing on semi synthetic opioid Buprenorphine?'' may require a search engine to be aware of several facts, including that Buprenorphine is a drug, that users refer to Buprenorphine with synonyms such as `bupe', `bupey', `suboxone', and `subbies', and the prescribed daily dosage range for Buprenorphine. The search engine should also have access to ontological knowledge as well as other ``intelligible constructs'' that are not typically modeled in ontologies, such as equivalent references to the frequency of drug use, the interval of use, and the typical dosage, to answer such complex search needs. At Kno.e.sis, we have developed an information retrieval system that integrates ontology-driven query interpretation with synonym-based query expansion and domain-specific rules to facilitate analysis of online web forums for drug abuse-related information extraction. Our system is based on a context-free grammar (CFG) that defines the interpretation of the query language constructs used to search for the drug abuse-related information needs and a domain-specific knowledge base that can be used to understand information in drug-related web forum posts. Our tool utilizes lexical, lexico-ontological, ontological, and rule-based knowledge to understand the information needs behind complex search queries and uses that information to expand the queries for significantly higher recall and precision (see Figure \ref{fig:fbgusers})~\cite{cameron2014hybrid}. This research~\cite{daniulaityte2013just} resulted in an unexpected finding of abuse of over the counter drug, which led to a FDA warning\footnote{\url{http://bit.ly/k-FDA}}. \\

\noindent \textbf{Application 4: Understanding city traffic using sensor and textual observations}

With increased urbanization, understanding and controlling city traffic flow has become an important problem. Currently, there are over 1 billion cars on the road network, and there has been a 236\% increase in vehicular traffic from 1981 to 2001~\cite{anantharam2016understanding}. Given that road traffic is predicted to double by 2020, achieving zero traffic fatalities and reducing traffic delays are becoming pressing challenges, requiring deeper understanding of traffic events, and their consequences and interaction with traffic flow. Sensors deployed on road networks continuously relay important information about travel speed through certain road networks while citizen sensors (i.e., humans) share real-time information about traffic/road conditions on public social media streams such as Twitter. As humans, we know how to integrate information from these multimodal data sources: qualitative traffic event information to account for quantitative measured traffic flow (e.g., an accident reported in tweets can explain a slow-moving traffic nearby). However, current research on understanding city traffic dynamics either focuses only on sensory data or only on social media data but not both. Further, we use historical data to understand traffic patterns and exploit the complementary and corroborative nature of these multimodal data sources to provide comprehensive information about traffic.

\begin{figure*}
\centering
\includegraphics[scale=0.5]{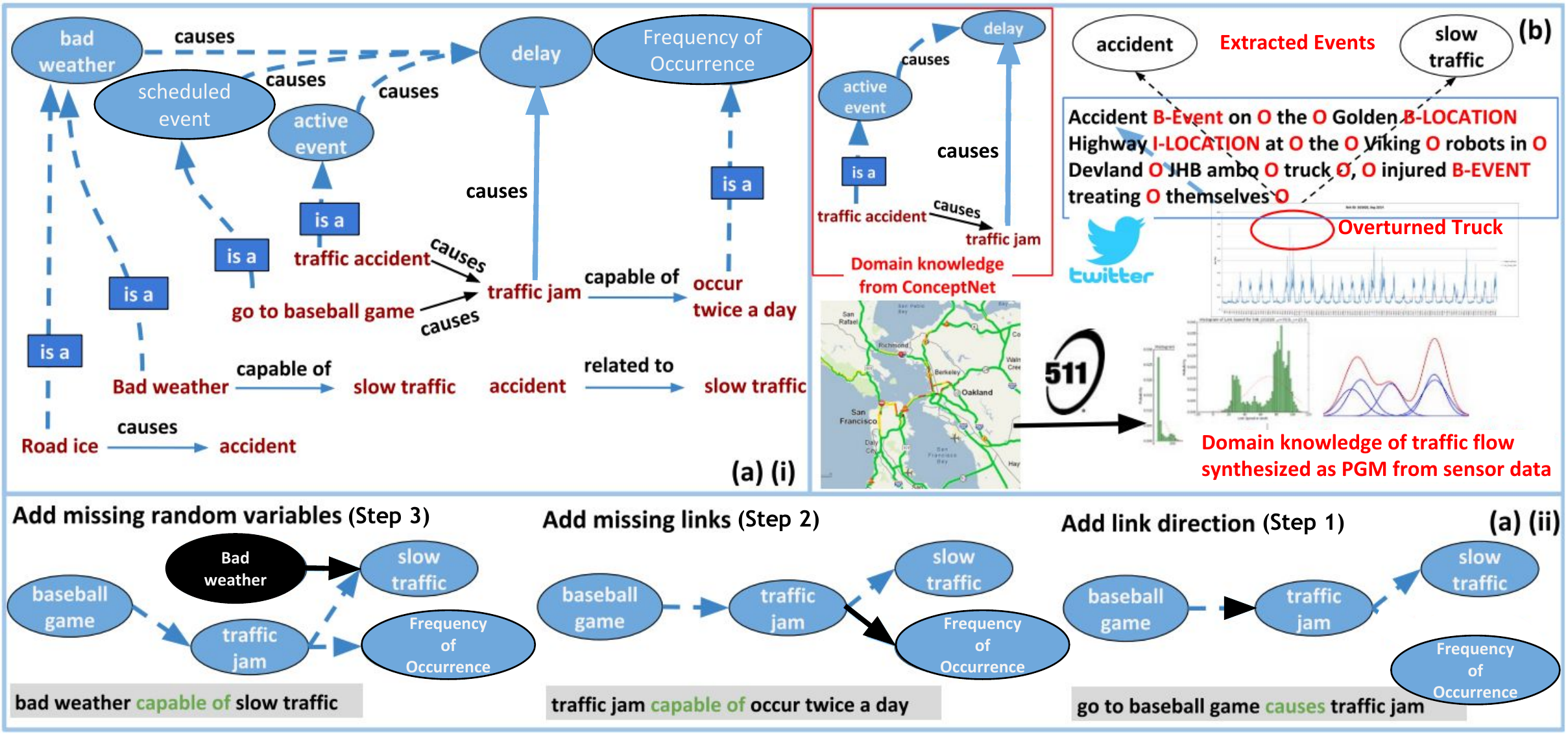}
\caption{(a)(i) Domain knowledge of traffic in the form of concepts and relationships (mostly causal) from the ConceptNet (a)(ii) Probabilistic graphical model (PGM) that explains the conditional dependencies between variables in traffic domain (only a portion is shown in the picture) is enriched by adding the missing random variables, links, and link directions extracted from ConceptNet. Figure~\ref{cityevents}(b) shows how this enriched PGM is used to correlate contextually related data of different modalities~\cite{anantharam2013traffic}.}
\label{cityevents}
\end{figure*}

One research direction is to create and materialize statistical domain knowledge about traffic into a machine-readable format. In other words, we want to define and establish associations between different variables (concepts) in the traffic domain (e.g., association between `bad weather' and a `traffic jam'). However, mining such correlations from data alone is neither complete nor reliable. We have developed statistical techniques based on probabilistic graphical models (PGMs)~\cite{koller2009probabilistic} to learn the structure (variable dependencies), leverage declarative domain knowledge to enrich and/or correct the gleaned structure due to limitations of a data-driven approach, and finally learn parameters for the updated structural model. Specifically, we use the sensor data collected by 511.org to develop an initial PGM that explains the conditional dependencies between variables in the traffic domain. Then we use declarative knowledge in ConceptNet to add/modify  variables (nodes) and the type and the nature of conditional dependencies (directed edges) before learning parameters, thereby obtaining the complete PGM. Figure~\ref{cityevents}(a)(i) shows a snippet of ConceptNet and Figure~\ref{cityevents}(a)(ii) demonstrates the enrichment step of the developed model using the domain knowledge in ConceptNet~\cite{anantharam2013traffic}.

Another research direction is to characterize a normal traffic pattern derived from sensor observations and then detect and explain any anomalies using social media data. We used a Restricted Switching Linear Dynamical System (RSLDS) to model normal speed and travel time dynamics and detect anomalies. Using speed and travel time data from each link, plus our common sense knowledge about the nature of expected traffic variations, we learn the parameters of the RSLDS model for each link. We then use a box-plot of the log likelihood scores of the various average speed traces with respect to the RSLDS model to learn and characterize anomalies for each link in the San Francisco Bay Area traffic data~\cite{anantharam2016understanding}. Later, given a new traffic speed trace over a link, we can obtain its log likelihood score with respect to the RSLDS model for the particular day of the week and the hour of the day, to determine whether it is normal or anomalous. This anomalous traffic speed information is further correlated with traffic events extracted from Twitter data (using crawlers seeded with OSM, 511.org and Scribe vocabularies) using their spatio-temporal context to explain the anomalies. Figure~\ref{cityevents}(b) demonstrates this process. \textbf{This example again demonstrates the vital role of multi-modal data for better interpretation of traffic dynamics, synthesizing probabilistic/statistical knowledge, and the application of both statistical models such as RSLDS and complementary semantic analysis of Twitter data}. Further exploration of different approaches to represent and exploit semantics appear in~\cite{sheth2005semantics}. Table~\ref{kbapplications} summarizes the role of knowledge bases in the four applications discussed above.

\begin{table}[]
\centering
\caption{Summary of knowledge-based approaches and the resulting improvements for each problem domain.}
\label{kbapplications}
\begin{tabular}{|l|l|l|}
\hline
\multicolumn{1}{|c|}{\textbf{Problem Domain}} & \multicolumn{1}{c|}{\textbf{Use of Knowledge bases}}          & \multicolumn{1}{c|}{\textbf{Nature of Improvement}} \\ \hline
Emoji Similarity and                          & Generation and application of                        & Leveraging linguistic knowledge                     \\
Sense Disambiguation                          & EmojiNet                                                            & for emoji interpretation                            \\ \hline
Implicit Entity Linking                       & Adapted UMLS definitions for       & Recall and coverage                                 \\
                       & identifying medical entities, and      &                                  \\
                       & Wikipedia and Twitter data for      &                                  \\
                       & identifying Twitter entities &                                                     \\ \hline
                                              
Understanding Drug              & Application of Drug Abuse                   & Recall and coverage                                 \\
Abuse-related                                   & Ontology  along with slang term                             &                                                     \\
Discussions                                  & dictionaries  and grammar                             &                                                     \\ \hline
Traffic Data Analysis                         & Statistical knowledge extraction                       & Anomaly detection and                  \\
                         & and using ontologies for Twitter                    &  explanation; Multi-modal data                \\
                                              &  event extraction                         & stream correlation                 \\ \hline
\end{tabular}
\end{table}

\section{Looking forward}
We discussed the importance of domain/world knowledge in understanding complex data in the real world, particularly when large amounts of training data are not readily available or it is expensive to generate. We demonstrated several applications where knowledge plays an indispensable role in understanding complex language constructs and multimodal data. Specifically, we have demonstrated how knowledge can be created to incorporate a new medium of communication (such as emoji), curated knowledge can be adapted to process implicit references (such as in implicit entity and relation linking),  statistical knowledge can be synthesized in terms of normalcy and anomaly and integrated with textual information (such as in traffic context), and linguistic knowledge can be used for more expressive querying of informal text with improved recall (such as in drug related posts). We are also seeing early efforts in making knowledge bases dynamic and evolve to account for the changes in the real world\footnote{\url{http://bit.ly/2cVGbov}}.

Knowledge seems to play a central role in human learning and intelligence, such as in learning from a small amount of data, and in cognition -- especially perception. Our ability to create or deploy just the right knowledge in our computing processes will improve machine intelligence, perhaps in a similar way as knowledge has played a central role in human intelligence. As a corollary to this, two specific advances we expect are: a deeper and nuanced understanding of content (including but not limited to text) and our ability to process and learn from multimodal data at a semantic level (given that concepts manifest very differently at the data level in different media or modalities). The human brain is extremely adept at processing multimodal data -- our senses are capable of receiving 11 million bits per second, and our brain is able to distill that into abstractions that need only a few tens of bits to represent (for further explorations, see~\cite{sheth2016semanticcognitive}). Knowledge plays a central role in this abstraction and reasoning process known as {\it the perception cycle\/}.

Knowledge-driven processing can be viewed from three increasingly sophisticated computational approaches: (1) Semantic Computing, (2) Cognitive Computing, and (3) Perceptual Computing. {\it Semantic Computing\/} refers to computing the type of a data value, and relating it to other domain concepts. In the healthcare context, this can involve relating symptoms to diseases and treatments. Ontologies, and Semantic Web technologies provide the foundation for semantic computing. {\it Cognitive computing\/} refers to representation and reasoning with data using background knowledge reflecting how humans interpret and process data. In the healthcare context, this requires capturing the experience and domain expertise of doctors through knowledge bases and heuristic rules for abstracting multimodal data into medically relevant abstractions, insights, and actions, taking into account triggers, personal data, patient health history, demographics data, health objectives, and medical domain knowledge. For instance, ``normal’’ blood pressure varies with factors such as age, gender, emotional state, activity, and illness; similarly, the ``target'' blood pressure, HBA1C, and cholesterol values a patient is advised to maintain depend on whether the patient is diabetic or not. In the traffic context, this can be used to interpret and label a time-series of traffic sensor data using a traffic event ontology. {\it Perceptual computing\/}, which builds on background knowledge created for semantic and cognitive computing, uses deductive reasoning to predict effects and treatments from causes, and abductive reasoning to explain the effects using causes, resolving any data incompleteness or ambiguity by seeking additional data. The knowledge itself can be a hybrid of deterministic and probabilistic rules, modeling both normalcy and anomalies, transcending abstraction levels. This directly contributes to making decisions and taking actions.

\begin{figure*}
\centering
\includegraphics[scale=0.37]{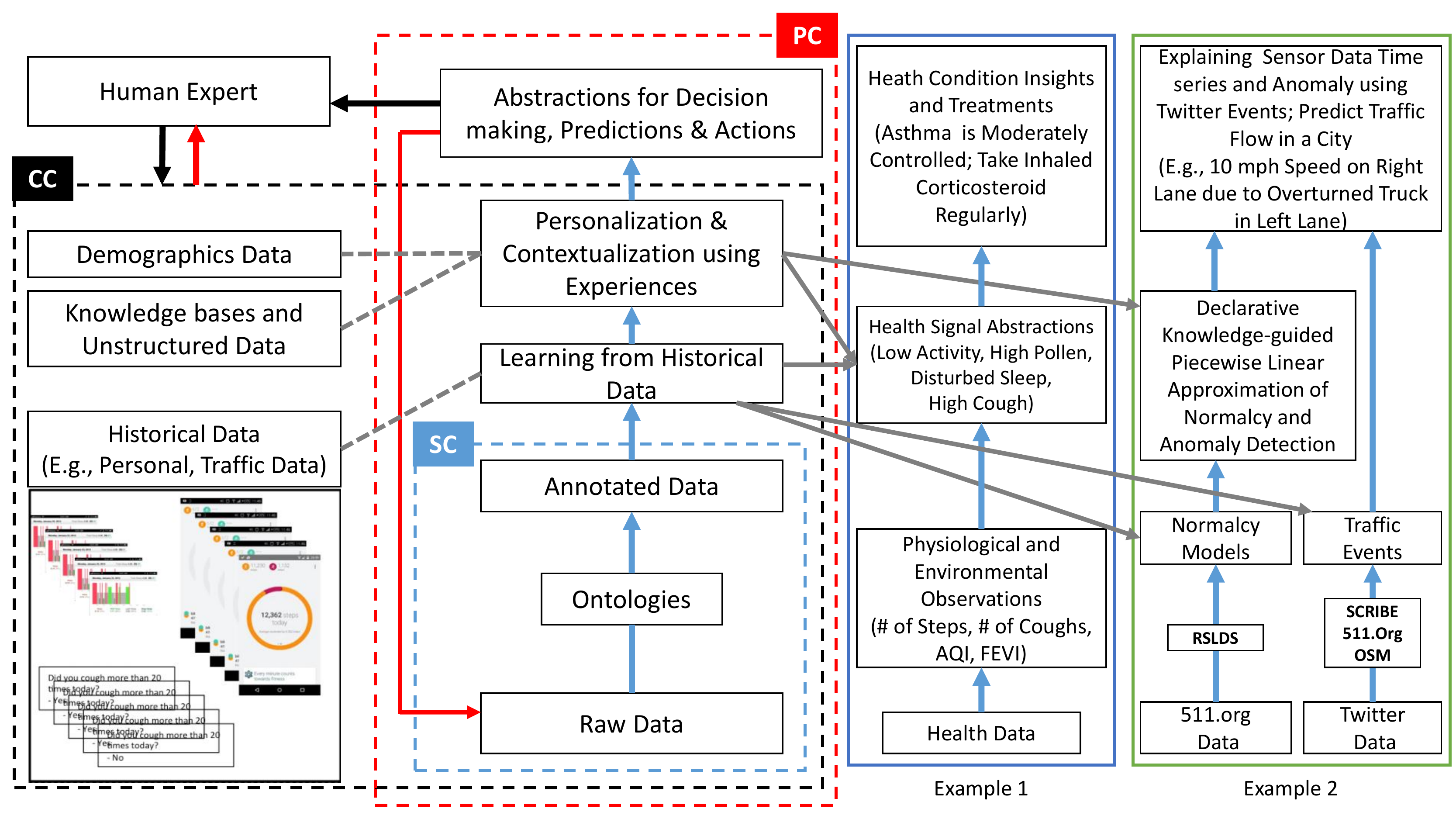}
\caption{Interplay between Semantic, Cognitive, and Perceptual Computing (SC, CC and PC) with Examples.}
\end{figure*}

We expect more progress in hybrid knowledge representation and reasoning techniques to better fit domain characteristics and applications. Even though deep learning techniques have made incredible progress in machine learning and prediction tasks, they are still uninterpretable and prone to devious attacks. There are anecdotal examples of misinterpretations of audio and video data through adversarial attacks that can result in egregious errors with serious negative consequences. In such scenarios, we expect hybrid knowledge bases to provide a complementary foundation for reliable reasoning. In the medical domain, the use of interleaved abductive and deductive reasoning (a.k.a., perception cycle) can provide actionable insights ranging from determining confirmatory laboratory tests and disease diagnosis to treatment decisions. Declarative medical knowledge bases can be used to verify the consistency of an EMR and data-driven techniques can be applied to a collection of EMRs to determine and fix potential gaps in the knowledge bases. Thus, there is a symbiotic relationship between the application of knowledge and data to improve the reliability of each other. The traffic scenario shows how to hybridize complementary statistical knowledge and declarative knowledge to obtain an enriched representation (See also~\cite{sheth2005semantics}). It also shows how multimodal data streams can be integrated to provide more comprehensive situational awareness.

Machine intelligence has been the holy grail of a lot of AI research lately. The statistical pattern matching approach and learning from big data, typically of a single modality, has seen tremendous success. For those of us who have pursued brain-inspired computing approaches, we think the time has come for rapid progress using a model-building approach. The ability to build broad models (both in terms of coverage as well as variety -- not only with entities and relationships but also representing  emotions, intentions and subjectivity features, such as, linguistic, cultural, and other aspects of human interest and functions)  will be critical. Further, domain-specific, purpose-specific, personalized declarative knowledge combined with richer representation -- especially probabilistic graph models -- will see rapid progress. These will complement neural network approaches. We may also see knowledge playing a significant role in enhancing deep learning. Rather than the dominance of data-centric approaches, we will see an interleaving and interplay of the data and knowledge tracks, each with its own strengths and weaknesses, and their combinations performing better than the parts in isolation.

\section*{Acknowledgments}

We acknowledge partial support from the National Institutes of Health (NIH) award: 1R01HD087132-01: ``kHealth: Semantic Multisensory Mobile Approach to Personalized Asthma Care'' and the National Science Foundation (NSF) award: EAR 1520870: ``Hazards SEES: Social and Physical Sensing Enabled Decision Support for Disaster Management and Response''. Points of view or opinions in this document are those of the authors and do not necessarily represent the official position or policies of the NIH or NSF.

\bibliographystyle{splncs03}

\end{document}